# Assessing the risk of re-identification arising from an attack on anonymised data

Anna Antoniou[1,a], Giacomo Dossena[1], Julia MacMillan[1], Steven Hamblin[1], David Clifton[1,2], Paula Petrone[1]

[1] Sensyne Health plc, Schrödinger Building, Heatley Road, Oxford Science Park, Oxford, OX44GE

[2] Department of Engineering Science, University of Oxford, Oxford, United Kingdom

[a] Corresponding author: Anna Antoniou, anna.antoniou@sensynehealth.com

**Objective:** The use of routinely-acquired medical data for research purposes requires the protection of patient confidentiality via data anonymisation. The objective of this work is to calculate the risk of re-identification arising from a malicious attack to an anonymised dataset, as described below. **Methods:** We first present an analytical means of estimating the probability of re-identification of a single patient in a $k$-anonymised dataset of Electronic Health Record (EHR) data. Second, we generalize this solution to obtain the probability of multiple patients being re-identified. We provide synthetic validation via Monte Carlo simulations to illustrate the accuracy of the estimates obtained. **Results:** The proposed analytical framework for risk estimation provides re-identification probabilities that are in agreement with those provided by simulation in a number of scenarios. Our work is limited by conservative assumptions which inflate the re-identification probability. **Discussion:** Our estimates show that the re-identification probability increases with the proportion of the dataset maliciously obtained and that it has an inverse relationship with the *equivalence* class size. Our recursive approach extends the applicability domain to the general case of a multi-patient re-identification attack in an arbitrary $k$-anonymisation scheme. **Conclusion:** We prescribe a systematic way to parametrize the $k$-anonymisation process based on a pre-determined re-identification probability. We observed that the benefits of a reduced re-identification risk that come with increasing $k$-size may not be worth the reduction in data granularity when one is considering benchmarking the re-identification probability on the size of the portion of the dataset maliciously obtained by the adversary.

# 1. INTRODUCTION

Aimed at improving the healthcare quality, Electronic Health Record (EHR) systems have been widely adopted to collect and store patient data, such as patient demographics, diagnosis codes, lab results and prescriptions, during normal clinical care [1-2]. However, due to their potential for knowledge discovery [3-4], such datasets are increasingly used for biomedical research [5-8]. Harnessing the knowledge-discovery potential of EHR databases legally and ethically requires the strict protection of patient confidentiality and data security. This is one of the main challenges facing EHR custodians as it requires close collaboration between policy makers, industry, and hospitals in order to ensure high-quality data collection under strict rules governing the preservation of patient confidentiality [9]. The ethical use of data is a topic of societal concern, illustrated by the recent media discussion regarding the sharing of UK NHS data with Google DeepMind [10] and the alleged use of Facebook data by Cambridge Analytica [11].

Such datasets need to be anonymised using state-of-the art algorithms that rely on minimising the probability of re-identification [12-13]. Modern laws such as the "European General Data Protection Regulation" of 2016 (GDPR) [14] and the "Health Insurance Portability and Accountability Act" of 1996 (HIPAA) [15] are concerned with the protection of the privacy of the individual. In the US and Canada, medical data released to a highly trusted recipient has a maximum threshold for re-identification of a single individual to 0.33 [16]. The rules are more stringent when the dataset is publicly released, confining the probability between 0.09 and 0.05 [16]. Given the operational guidelines established for the anonymisation of health data [17-18], it is critical to put in place anonymisation techniques for which the probability of re-identification is accurately estimated.

$K$-anonymisation is a privacy-preserving algorithm that has been used to anonymise EHR databases prior to release [19-21]. It relies on grouping data from similar patients into "equivalence classes" composed of $k$ members such that the records associated with individual patients are indistinguishable. The resulting probability of re-identification of a single patient, provided the hypothesised intruder has prior information such as publicly available data that can be used to identify the patient, and access to the entire anonymised dataset, is at worst $1/k$ [22].

Here we derive an analytical solution to quantify an estimate of re-identification probability of a single patient in a $k$-anonymised dataset and extend that to obtain the probability of multiple patients being re-identified in an attack. Taken as a whole, our work provides a systematic way to parametrise $k$-anonymisation based on the target re-identification probability that can be tolerated for a specific dataset.

## 1.1. Introduction to $k$-anonymity

Re-identification attacks involve three kinds of data attributes: (i) *direct identifiers*, (ii) *quasi-identifiers* and (iii) *sensitive attributes*. Any information that directly identifies individuals via a one-to-one mapping (such as a national health insurance number) is a *direct identifier* (Table 1). Attributes not directly identifying an individual but which can do so when used in combination with other data are *quasi-identifiers* (e.g., sex, age, postcode, ethnicity [34]). *Sensitive attributes* include all "sensitive" information, such as medical diagnoses [35].

| Patient NHS number | Sex | Age | Postcode | Ethnicity | Length of stay (days) | Primary diagnosis |
|---|---|---|---|---|---|---|
| 123 456 7890 | M | 50 | OX12 6HU | White | 2 | Cancer |
| 123 456 7890 | M | 50 | OX12 6HU | White | 6 | Pneumonia |
| 111 342 9807 | M | 53 | OX12 9HU | White | 3 | Cancer |
| 111 342 9807 | M | 53 | OX12 9HU | White | 11 | Stroke |
| 878 452 0908 | M | 55 | OX12 8JU | White | 4 | Pneumonia |
| 878 452 0908 | M | 55 | OX12 8JU | White | 7 | Influenza |
| 878 452 0908 | M | 55 | OX12 8JU | White | 16 | Pneumonia |
| 673 542 8745 | F | 64 | OX14 6GU | Mixed | 22 | Cancer |
| 879 543 8132 | F | 65 | OX14 7GU | Mixed | 24 | Cancer |
| 763 356 4625 | F | 65 | OX14 8GU | Mixed | 27 | Cancer |
| 989 323 3221 | F | 58 | OX13 6AB | Mixed | 4 | Influenza |
| 837 473 7584 | F | 62 | OX13 6AF | Mixed | 2 | Influenza |
| 878 462 9834 | F | 60 | OX13 6BX | Mixed | 3 | Ulcer |
| 878 663 2210 | F | 56 | OX13 6FW | White | 4 | Stroke |

*Table 1: Simplistic example of a dataset to be anonymised. This artificial dataset is based on an extremely basic Patient Admission System and it is useful for illustrative purposes. Exposition to follow does not assume such a simple data structure. All patient attributes are fictitious.*

In an anonymisation process, all direct identifiers are replaced by a unique and randomized number. During a re-identification attack, we assume that an adversary has access to some commonly-available public data (Table 2) and that they will operate upon the anonymised dataset (Table 1) in an attempt to re-identify patients. For example, even though the name "John Doe" does not appear in the target dataset (Table 1), if the adversary knows that this individual was hospitalised, they can derive the reasons of hospitalisation carrying out a successful identity attack.

| Name | Sex | Age | Postcode | Ethnicity |
|------|-----|-----|----------|-----------|
| John Doe | M | 50 | OX12 6HU | White |

*Table 2: Item of assumedly-public information available to an adversary in our exemplar*

$K$-anonymisation is based on generalisations and suppressions of quasi-identifiers. Given a positive integer $k$, the algorithm will generalize the quasi-identifiers (e.g., by removing the last three digits from the postcode) and group patients into sets of $k$ members or more sharing the same quasi-identifiers such that they are indistinguishable [20,34]. These groups are said to form equivalence classes. In our example by setting $k = 3$, we anonymise our simplistic example (Table 2), resulting in three distinct equivalence classes (C1, C2, and C3) seen in Table 3. Equivalence classes C1 and C2 contain three patients, satisfying the minimum requirement for k-anonymisation, whereas the third equivalent class C3 exceeds that criterion. Note that this is one of many possible groupings resulting in equivalent classes (e.g., a higher degree of quasi-identifier generalisation can result in bigger equivalence class sizes at a cost of higher information loss). This procedure limits the re-identification probability to 1/3 at most.

| Patient ID | Sex | Age | Postcode | Ethnicity | Length of stay (days) | Diagnosis | Equivalence group label |
|---|---|---|---|---|---|---|---|
| 2887 | M | 50-55 | OX12 | White | 0-5 | Cancer | |
| 2887 | M | 50-55 | OX12 | White | 6-15 | Pneumonia | |
| 3679 | M | 50-55 | OX12 | White | 0-5 | Cancer | |
| 3679 | M | 50-55 | OX12 | White | 6-15 | Stroke | C1 |
| 1208 | M | 50-55 | OX12 | White | 0-5 | Pneumonia | |
| 1208 | M | 50-55 | OX12 | White | 6-15 | Influenza | |
| 2257 | F | 60-65 | OX14 | Mixed | 20-25 | Cancer | |
| 9006 | F | 60-65 | OX14 | Mixed | 20-25 | Cancer | C2 |
| 8773 | F | 60-65 | OX14 | Mixed | 20-25 | Cancer | |
| 4653 | F | 55-65 | OX13 | Mixed | 0-5 | Influenza | |
| 7363 | F | 55-65 | OX13 | Mixed | 0-5 | Influenza | |
| 5392 | F | 55-65 | OX13 | Mixed | 0-5 | Ulcer | C3 |
| 6453 | F | 55-65 | OX13 | Mixed | 0-5 | Stroke | |

*Table 3: k-anonymised data. Each equivalence class contains at least three patients with any number of records each.*

## 2. METHODS

### 2.1. Predicting the risk of re-identification of a single patient

In this section we derive a probabilistic model to estimate the risk of re-identification of a single individual. We note that the applicability of this method extends beyond healthcare data to any domain in which records from individuals have been anonymised using $k$-anonymisation. First, we establish assumptions that delimit the range of applicability of our model.

#### 2.1.1. Assumptions and notations

We define our dataset $\mathcal{D}$ to be an $n \times m$ matrix consisting of information from $D$ individuals having been $k$-anonymised. This creates a set of equivalence classes $\{b_i^{\mathcal{D}} : i \in \mathbb{N}\}$ such that $\mathcal{D} = \bigcup_i b_i^{\mathcal{D}}$. Next, we define $X$ to be an arbitrary patient who is the target of a re-identification attack whose records exists within the anonymised dataset $\mathcal{D}$. The medical history of patient $X$ consists of a series of records each corresponding to a hospital admission.

We assume that the adversary knows $X$ is in $\mathcal{D}$ and knows the value of each quasi-identifier of every admission record of $X$ but does not know any of the sensitive information. For example, if the adversary seeks to re-identify John Doe (Table 2), they know the sex, ethnicity, postcode and length of stay for any one or all of his hospital admissions but does not know that the patient has been diagnosed with cancer and pneumonia. As such, we can assume that the adversary has access to a public dataset containing all the relevant quasi-identifiers that will be used for a record linkage attack; each patient in the anonymised dataset is assumed to exist and is unique in the external dataset held by the adversary.

We further define a maliciously obtained leaked dataset $\mathcal{L} \subseteq \mathcal{D}$ consisting of $L$ patients and a set of equivalence classes $\{b_i^{\mathcal{L}}: i \in \mathbb{N}\}$ such that $\mathcal{L} = \bigcup_i b_i^{\mathcal{L}}$. Note that $b_i^{\mathcal{L}} \subseteq b_i^{\mathcal{D}}$ and therefore $b_i^{\mathcal{L}}$ might be partially full or empty, as not all patients will be leaked. $\mathcal{L}$ contains only entire patient histories, not a selection of records from a patient (e.g., all of John Doe's records are present in their entirety, or none at all). Lastly, for a given attack, every leaked subset arising from any combination of L whole patient histories is assumed to be equiprobable, as every patient has an equal probability of being part of the leaked dataset. For instance, if $\mathcal{D}$ occurs as in Table 3 then a possible leaked dataset $\mathcal{L}$ is seen in Table 4. Here $D = 10$ and $L = 3$. Note that $\mathcal{L}$ contains two patients from C1, and one patient from C3.

| Patient id | sex | Age   | Postcode | Ethnicity | LOS  | Diagnosis |
|------------|-----|-------|----------|-----------|------|-----------|
| 3679       | M   | 50-55 | OX12     | White     | 0-5  | Cancer    |
| 3679       | M   | 50-55 | OX12     | White     | 6-15 | Stroke    |
| 1208       | M   | 50-55 | OX12     | White     | 0-5  | Pneumonia |
| 1208       | M   | 50-55 | OX12     | White     | 6-15 | Influenza |
| 4653       | F   | 55-65 | OX13     | Mixed     | 0-5  | Influenza |

*Table 4: Example of a leaked dataset based on Table 3*

## 2.1.2. Estimating the probability of re-identification

Here, in order to estimate the upper bound of the re-identification probability and for ease of calculation, we further assume each equivalence class in $\mathcal{D}$ has exactly $k$ patients., i.e., $|b_i^{\mathcal{D}}| = k \ \forall \ i$.

A re-identification attack, as described here, can be characterised by $D$, $L$ and $k$. We are then led to consider the following list of events within our sample space:

- $\mathcal{E}_i^{\mathcal{D}} : X \in b_i^{\mathcal{D}}$,

- $\mathcal{E}_i^{\mathcal{L}} : X \in b_i^{\mathcal{L}}$,

- $\mathcal{E}_{i,h}^{\mathcal{L}} : |b_i^{\mathcal{L}}| = h$, where $|A|$ denotes the cardinality of set $A$,

- $\mathcal{E}_I$: patient $X$ is re-identified by the adversary.

Our goal is to find $P(\mathcal{E}_I)$ for a single patient. By definition $x \in \mathcal{E}_i^{\mathcal{L}} \implies x \in \mathcal{E}_i^{\mathcal{D}}$ and therefore $P(\mathcal{E}_i^{\mathcal{L}}) = P(\mathcal{E}_i^{\mathcal{L}} \cap \mathcal{E}_i^{\mathcal{D}}) = P(\mathcal{E}_i^{\mathcal{L}}|\mathcal{E}_i^{\mathcal{D}})P(\mathcal{E}_i^{\mathcal{D}})$. We further partition event $\mathcal{E}_i^{\mathcal{L}}$ to events $\mathcal{E}_{i,h}^{\mathcal{L}}$. Thus, $\mathcal{E}_i^{\mathcal{D}} \cap \mathcal{E}_i^{\mathcal{L}} \cap \mathcal{E}_{i,h}^{\mathcal{L}}$ and $\mathcal{E}_j^{\mathcal{D}} \cap \mathcal{E}_j^{\mathcal{L}} \cap \mathcal{E}_{j,h}^{\mathcal{L}}$ are mutually exclusive whenever $i \neq j$ and partition the set of all outcomes where a person is re-identified. Additionally, the pairs of events $\mathcal{E}_i^{\mathcal{L}}, \mathcal{E}_j^{\mathcal{L}}$ and $\mathcal{E}_i^{\mathcal{D}}, \mathcal{E}_j^{\mathcal{D}}$ respectively are mutually exclusive whenever $i \neq j$. Finally, we have that $\mathcal{E}_{i,h}^{\mathcal{L}}$ and $\mathcal{E}_{i,h'}^{\mathcal{L}}$ are mutually exclusive whenever $h \neq h'$ since, given a leak, the size of the $i^{\text{th}}$ equivalence class in the leaked dataset can only have one value. We can then write $P(\mathcal{E}_I)$ as a sum of probabilities, as follows:

$$P(\mathcal{E}_I) = \sum_{i=1}^{D/k} \sum_{h=1}^{k} P\left(\mathcal{E}_I \cap \mathcal{E}_i^{\mathcal{L}} \cap \mathcal{E}_i^{\mathcal{D}} \cap \mathcal{E}_{i,h}^{\mathcal{L}}\right). \quad (1)$$

where the inner and outer summations span the varying equivalence class sizes in $\mathcal{L}$ and the equivalence class sizes in $\mathcal{D}$ respectively. Since all equivalence classes in $\mathcal{D}$ have the same size $|b_i^{\mathcal{D}}| = k$, all the summands in the double sum above do not depend on $i$ and thus:

$$P(\mathcal{E}_I) = \frac{D}{k} \sum_{h=1}^{k} P\left(\mathcal{E}_I \cap \mathcal{E}^{\mathcal{L}} \cap \mathcal{E}^{\mathcal{D}} \cap \mathcal{E}_h^{\mathcal{L}}\right) \quad (2)$$

where $\mathcal{E}^{\mathcal{L}} = \mathcal{E}_i^{\mathcal{L}}$, $\mathcal{E}^{\mathcal{D}} = \mathcal{E}_i^{\mathcal{D}}$, $\mathcal{E}_h^{\mathcal{L}} = \mathcal{E}_{i,h}^{\mathcal{L}}$ with $i = 1$. By applying the Bayesian chain rule to *Equation (2)* we obtain:

$$\begin{aligned} P(\mathcal{E}_I) = \frac{D}{k} \sum_{h=1}^{k} \quad & P(\mathcal{E}_I \mid \mathcal{E}^{\mathcal{L}} \cap \mathcal{E}^{\mathcal{D}} \cap \mathcal{E}_h^{\mathcal{L}}) \\ \times \ & P(\mathcal{E}^{\mathcal{L}} \mid \mathcal{E}^{\mathcal{D}} \cap \mathcal{E}_h^{\mathcal{L}}) \\ \times \ & P(\mathcal{E}^{\mathcal{D}}) \times P(\mathcal{E}_h^{\mathcal{L}}). \end{aligned} \quad (3)$$

It is now straightforward to assign a value to each of the factors in *Equation (3)*:

- $P(\mathcal{E}_I \mid \mathcal{E}^{\mathcal{L}} \cap \mathcal{E}^{\mathcal{D}} \cap \mathcal{E}_h^{\mathcal{L}})$ is the probability of identifying $X$, knowing that belongs to the equivalence class $b^{\mathcal{L}} \subseteq b^{\mathcal{D}}$ of size $h$ in the leaked dataset. This is given by $\frac{1}{h}$.

- $P(\mathcal{E}^{\mathcal{L}} \mid \mathcal{E}^{\mathcal{D}} \cap \mathcal{E}_h^{\mathcal{L}})$ is the probability of $X$ being in the equivalence class $b^{\mathcal{L}}$, knowing that that equivalence class is of size $h$, and is given by $\frac{h}{k}$.

- $P(\mathcal{E}^{\mathcal{D}})$ is the probability of $X$ being in the equivalence class $b^{\mathcal{D}}$, and is given by $\frac{k}{D}$.

- $P(\mathcal{E}_h^{\mathcal{L}})$ is the probability that the equivalence class $b^{\mathcal{L}}$ is of size $h$. To find this probability we reason as follows: the number of all possible leaks of size $L$ is given by the combinatorial number $\binom{D}{L}$. Next, we count how many of these leaks correspond to a size $h$ for equivalence class $b^{\mathcal{L}}$. The number of ways to select a subset of $b^{\mathcal{D}}$ of size $h$ is $\binom{k}{h}$, and for each of these we have $\binom{D-k}{L-h}$ many ways of arranging the remaining leaked patients outside of $b^{\mathcal{D}}$. This gives:

$$P(\mathcal{E}_h^{\mathcal{L}}) = \frac{\binom{k}{h}\binom{D-k}{L-h}}{\binom{D}{L}} . \tag{4}$$

Combining the above, we obtain:

$$\begin{aligned} P(\mathcal{E}_I) &= \frac{D}{k}\sum_{h=1}^{k}\left[\frac{1}{h}\times\frac{h}{k}\times\frac{k}{D}\times\frac{\binom{k}{h}\binom{D-k}{L-h}}{\binom{D}{L}}\right] \\ &= \frac{1}{k}\sum_{h=1}^{k}\frac{\binom{k}{h}\binom{D-k}{L-h}}{\binom{D}{L}} . \end{aligned} \tag{5}$$

Eq. (5) is the final theoretical solution for the probability of re-identifying $X$ given all of the above assumptions.

## 2.2 Simulating a re-identification attack

We then evaluate the probabilities of patient re-identification for varying leak $L$ and equivalence class size $k$ by simulating an identity attack, following the steps below:

1. Construct a leak set $L$ by randomly sampling from the anonymised dataset $\mathcal{D}$.

2. Assign a risk value $r$ to each patient representing the probability of identification such that:

$$r := \begin{cases} 0 & X \notin \mathcal{L} \\ \frac{1}{|b_i^{\mathcal{L}}|} & X \in \mathcal{L} \text{ and } X \in b_i^{\mathcal{L}} \end{cases} \tag{6}$$

3. The average risk is computed over all patients in the leaked set.

4. Repeat steps 1-3 several times to obtain the distribution of the mean risk value over all patients and the 95% confidence interval. The probability of re-identification is taken as the expected value of the average risk for the specific leak and equivalence class size

## 2.3 Recursive solution to the probability of multiple patient re-identifications

We next extend our solution to a more complex problem: the re-identification of multiple individuals in one attack, by providing a recursive algorithm which further allows a more general scenario in which there is a distribution of equivalence classes, as it is the case in standard anonymisation procedures.

### 2.3.1. Recursive re-identification example

We assume a scenario in which we wish to estimate the probability of the event that three patients $(A,B,C)$ are re-identified $\mathcal{E}_{I_{A:C}}$ from a $k$-anonymised dataset where $k \leq |b_i^\mathcal{D}| \leq K$. The initial state of the system can be described by:

- $n$: the number of patients to be re-identified

- $D$: the size of the dataset

- $L$: the size of the leaked dataset

- **$a$:** array where the $i^{th}$ element is defined as $a_i := |\{\ell: |b_\ell^\mathcal{D}| = i\}|$ where $0 \leq i \leq K$

The probability of re-identifying the first patient $A$ depends on the size of the equivalence class they come from and on the number of other patients $j \in \{0, ..., k-1\}$ from the same equivalence class that are also part of the leaked dataset, such that:

$$P(\mathcal{E}_{I_A}) = \sum_{k=\min|b^\mathcal{D}|}^{K} \sum_{j=0}^{k-1} P(\mathcal{E}_k^j) \qquad (7)$$

where $\mathcal{E}_k^j$ is the event that $A$ is re-identified from $b_i^\mathcal{L}$ with $|b_i^\mathcal{D}| = k$ and $|b_i^\mathcal{L}| = j+1$, i.e., $j$ other patients are leaked from the same equivalence class as $A$. Equivalently, as seen in Section 2.1:

$$\begin{aligned}
P(\mathcal{E}_k^j) &= P(\mathcal{E}_{I_A} \cap (A \in \mathcal{L}) \cap (A \in b_i^\mathcal{D}: |b_i^\mathcal{D}| = k) \cap (A \in b_i^\mathcal{L}: |b_i^\mathcal{L}| = j+1) \\
&= P(\mathcal{E}_{I_A} | (A \in \mathcal{L}) \cap (A \in b_i^\mathcal{D}: |b_i^\mathcal{D}| = k) \cap (A \in b_i^\mathcal{L}: |b_i^\mathcal{L}| = j+1) \\
&\quad \times P\left((A \in b_i^\mathcal{L}: |b_i^\mathcal{L}| = j+1) | (A \in b_i^\mathcal{D}: |b_i^\mathcal{D}| = k)\right) \\
&\quad \times P(A \in b_i^\mathcal{D}: |b_i^\mathcal{D}| = k) \\
&\quad \times P(A \in \mathcal{L})
\end{aligned} \qquad (8)$$

For the general case of re-identifying patient A where the equivalence class sizes $k$ in $\mathcal{D}$ are $k = \{1, ..., K\}$, the sample space consists of the following events:

$$\begin{aligned}
k = 1: &\quad \mathcal{E}_1^0 \\
k = 2: &\quad \mathcal{E}_2^0, \quad \mathcal{E}_2^1 \\
k = 3: &\quad \mathcal{E}_3^0, \quad \mathcal{E}_3^1, \quad \mathcal{E}_3^2 \\
&\quad \vdots \\
k = K: &\quad \mathcal{E}_K^0, \quad \mathcal{E}_K^1, \quad \mathcal{E}_K^2, \quad ... \quad \mathcal{E}_K^{K-1}.
\end{aligned} \qquad (9)$$

We assume that the equivalence class size $k = 2$. Then our initial state is:

$$L, D, n = 3 \text{ and } \boldsymbol{a} = [a_0, a_1, a_2] \text{ where } a_0 = a_1 = 0,$$

and the re-identification probability for our first patient $A$ is:

$$\begin{aligned}
P(\mathcal{E}_{I_A}) &= P(\mathcal{E}_2^0) + P(\mathcal{E}_2^1) \\
&= 1 \times \frac{L}{D} \times \frac{\binom{1}{0}\binom{D-2}{L-1}}{\binom{D-1}{L-1}} \times \frac{a_2}{a_1+a_2} \\
&\quad + \frac{1}{2} \times \frac{L}{D} \times \frac{\binom{1}{1}\binom{D-2}{L-2}}{\binom{D-1}{L-1}} \times \frac{a_2}{a_1+a_2}
\end{aligned} \qquad (10)$$

If we let the probability of re-identification of patient $X \in b_i^\mathcal{D}$ be $X_j$ where $j = |b_i^\mathcal{D}|$, then we can follow a recursive tree to find the probability of re-identifying a second patient $B$. Figure 1 denotes such a tree; the root node denotes the probability $A_2$ of re-identifying the first patient $A$ where the initial state contains only equivalent classes of size 2.

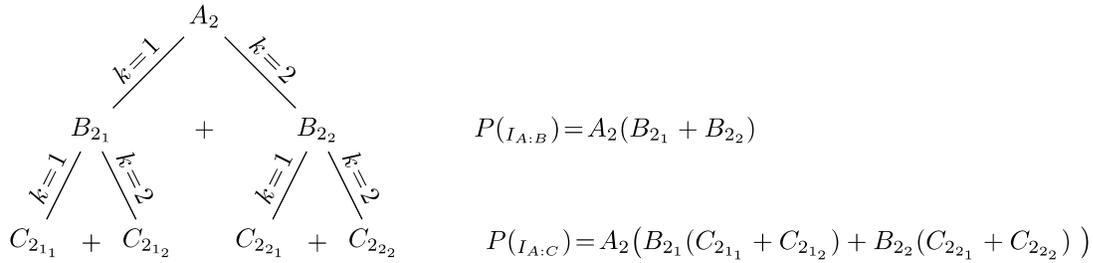

$$P(I_{A:B}) = A_2(B_{2_1} + B_{2_2})$$

$$P(I_{A:C}) = A_2\big(B_{2_1}(C_{2_{1_1}} + C_{2_{1_2}}) + B_{2_2}(C_{2_{2_1}} + C_{2_{2_2}})\big)$$

Figure 1: Recursive tree for re-identifying patients $A, B$ and $C$. The root node denotes the probability $A_2$ of re-identification of a single patient where the initial state contains only equivalent classes of size 2. To obtain the probability of re-identification, sibling nodes are added, and their summand is multiplied up with their parent node.

Now, because A has already been re-identified, the remaining leaked dataset will contain one equivalence class of size 1, and one less equivalence class of size 2 and the new updated state is:

$$L - 1, D - 1, n = 2, \text{ and } \mathbf{a} = [\,a_0,\, a_1 + 1,\, a_2 - 1\,]\text{ where } a_0 = 0 \text{ and } a_1 = 0.$$

B could now belong in $k = 1$ or in one of the $k = 2$ equivalence classes. We carry out the same calculations as above but with the new system state. $B_{2_1}$ is the probability of re-identification of the second patient B given they come from $k = 1$ and the previously re-identified patient A originated from $k = 2$. The subscripts denote the history of a patient: the first subscript denotes the equivalence class size of the previously re-identified patient and the second is the equivalence class size that the patient is in. $B_{2_1}$ and $B_{2_2}$ are additive and their sum represents the probability of identifying B given A has been re-identified:

$$P(\mathcal{E}_{I_B} \mid \mathcal{E}_{I_A}) = B_{2_1} + B_{2_2} \quad (11)$$

The probability of identifying both A and B is obtained by multiplying $B_{2_2} + B_{2_1}$ with its parent node $A_2$ such that:

$$\begin{aligned} P(\mathcal{E}_{I_B} \cap \mathcal{E}_{I_A}) &= P(\mathcal{E}_{I_A}) \times P(\mathcal{E}_{I_B} \mid \mathcal{E}_{I_A}) \\ &= A_2 \times (B_{2_1} + B_{2_2}) \end{aligned} \quad (12)$$

The same logic is followed for identifying the third patient C. In pseudo code, the probability of re-identifying n patients $P(\mathcal{E}_{I_{1:n}})$ is:

```
1: function Probability(a,L,D,n):
2:     P = 0
3:     for each element i of array a do:
4:         compute P($\mathcal{E}_{I_X}$) for patient X in a[i]
5:         update a as a' with $a'_{i-1} = a_{i-1} + 1$ and $a'_i = a_i + 1$
6:         P = P + P($\mathcal{E}_{I_X}$) × Probability(a', L – 1, D – 1, n – 1)
7:     end for
8:     return P
9: end function
```

We also simulate multiple patient re-identification iteratively, as showed by the following pseudo code for a single $L$:

```
1: P = 1
2: create random $\mathcal{L}$
3: for i ← 1,n do:
4:     randomly choose a patient X from $\mathcal{L}$
5:     read corresponding risk value r according to Equation (6)
6:     P = P × r
7:     remove patient X from $\mathcal{L}$ such that $\mathcal{L} = \mathcal{L}\setminus\{X\}$
8:     update the risk value for patients in the same equivalence class as X
9: return P
```

# 3. RESULTS

## 3.1 Single patient re-identification

Using *Equation (5)* in Figure 2 we compare the analytical (solid line) and simulated (solid dots with 95% CI) probability of re-identification with $D = 10000$ and $L$ in the range $1000 \leq L \leq 4000$ for varying values of $k$. The two approaches agree. As expected, the probability of re-identification decreases for increasing equivalence class size. Additionally, as the leak size approaches the total dataset size $D$, the probability of identification quickly tends to $\frac{1}{k}$ (horizontal dotted line). This could be attributed to the very conservative nature of our assumptions, specifically that the entire history of a patient is leaked rather than only a part of it. Observe that the impact of increasing $L$ on the re-identification probability is amplified for smaller equivalence class sizes, as seen by the increasing re-identification probability range for decreasing equivalence class size.

To characterise the simulated re-identification probability distribution (rather than the expected value), we plot the entire distribution obtained from the simulations for different values of $k$ and $L$. In Figure 3a we observe the simulated probability of re-identification (left) and the simulated risk distribution (right) for *D = 10000, L = 4000* and varying *k*. The relative frequency of the simulated risk is shown for *k = 1, 10* and *20.* The reduction in the risk with increasing $k$-size plateaus for $k > 10$, while the risk shifts from a binary outcome with peaks at 0 (patient $\in \mathcal{L}$) and 1 (patient $\notin \mathcal{L}$) for $k = 1$ to a distribution of values that are closer to zero. Figure 3b illustrates the probability of re-identification (left) and the simulated risk distribution (right) for *D = 10000, k = 5* and varying *L*. Similar shifts in the risk distribution are observed for increasing $L$; the shift away from zero signifies more patients falling into $\mathcal{L}$. As seen in both *Figures 3a* and *3b,* the density shows a prominent increase in the range (0,1) as more patients fall into the same $b_i^{\mathcal{L}}$ for increasing $k$ and $L$ respectively.

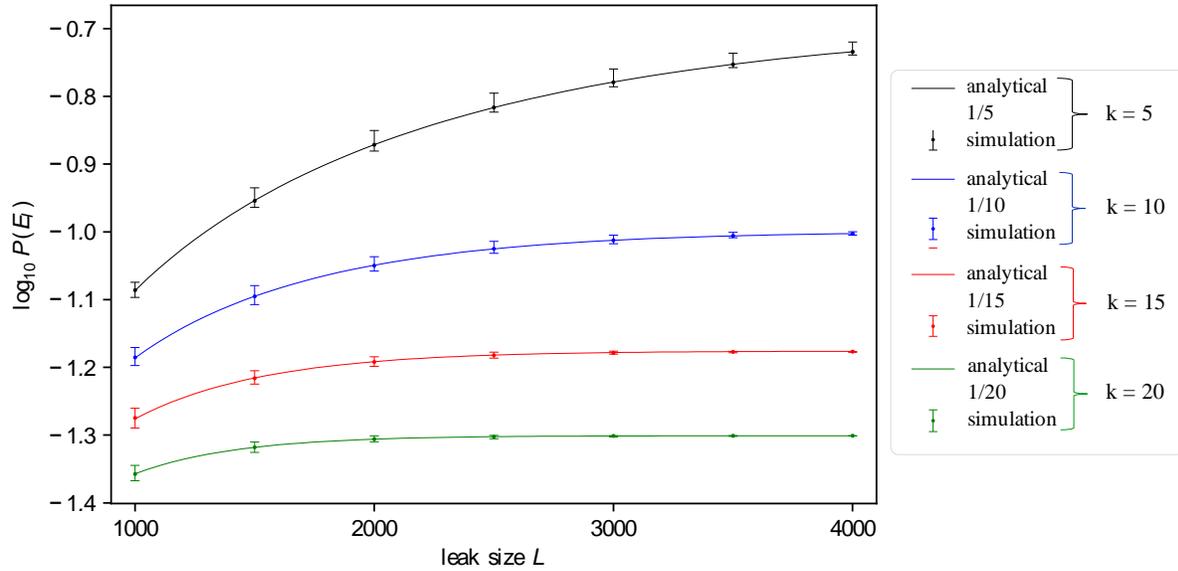

*Figure 2: Probability (log-scale) of re-identification with increasing L and varying k.*

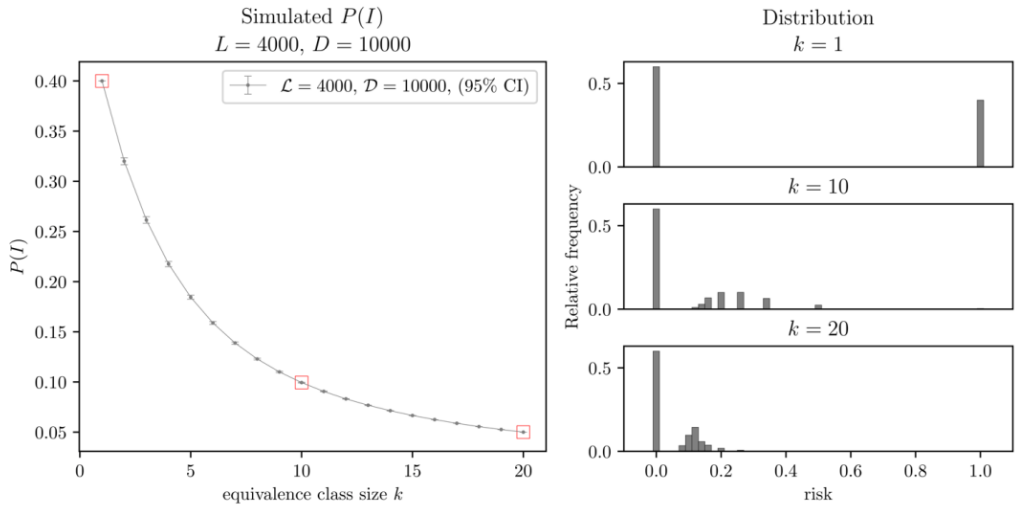

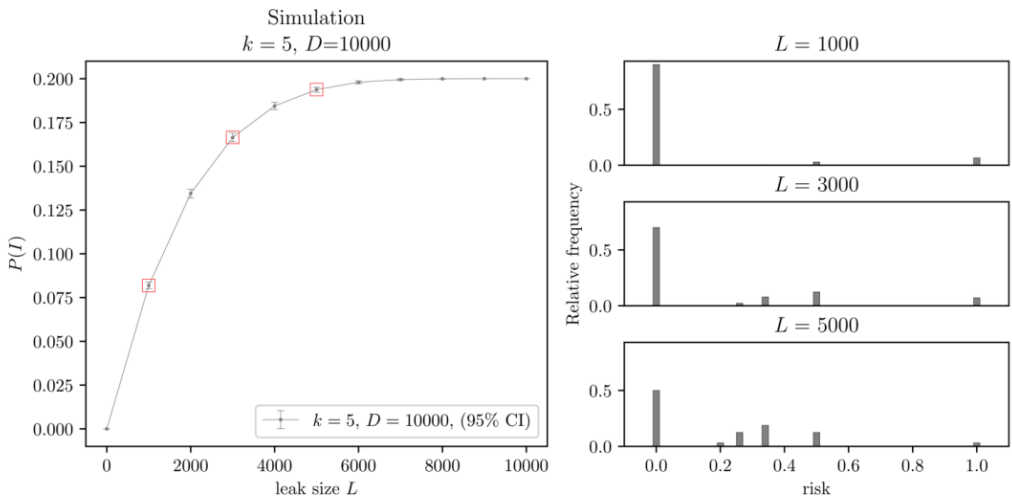

Figure 3: *Monte Carlo simulations of re-identification probabilities agree with analytical solution. (a) The simulated probability of re-identification (left) and the simulated risk distribution (right) for D = 10000 patients, leak size L = 4000 and varying k. The relative frequency of the simulated risk is shown for k = 1,10 and 20. (b) The probability of re-identification (left) and the simulated risk distribution (right) for D = 10000 patients, k = 5 and varying leak size L = 1000, 2000, 3000, 4000.*

## 3.2 Multiple patient re-identification results: recursive solution and simulation

The recursive solution (smooth line) agrees with the analytical solution (solid dots) for the case of re-identification of a single individual for homogeneous $k$ and varying $L$ (see Supplementary Material). The same can be said for the case of multi-patient re-identifications using both the recursive algorithm and Monte Carlo simulation where the entire dataset has been leaked. Figure 4$a$ illustrates the agreement of the recursive algorithm (solid line) with the simulation (solid dots) along with the decreasing trend of the re-identification probability with increasing $n$ for each equivalence class size. Figure 4$b$ shows the re-identification probability as a function of $L$. We see an increase in the re-identification probability for increasing $L$ regardless of $n$, and similar to Figure 3$a$ we observe that the re-identification probability decreases with increasing $n$.

In addition, we observe that the re-identification probabilities for multi-patient attacks using a fixed $k$-size forms the upper bound of the more general case of a heterogeneous equivalence class size distribution. *Figure 5* illustrates the re-identification probabilities for such an attack using both a homogeneous and multiple instances of a heterogeneous array $\boldsymbol{a}$. The latter are randomly drawn from a lower-tail truncated normal with both a mean and minimum of 5. The re-identification probability given a fixed $k$ seems to form an upper bound of the more general case of a heterogeneous state.

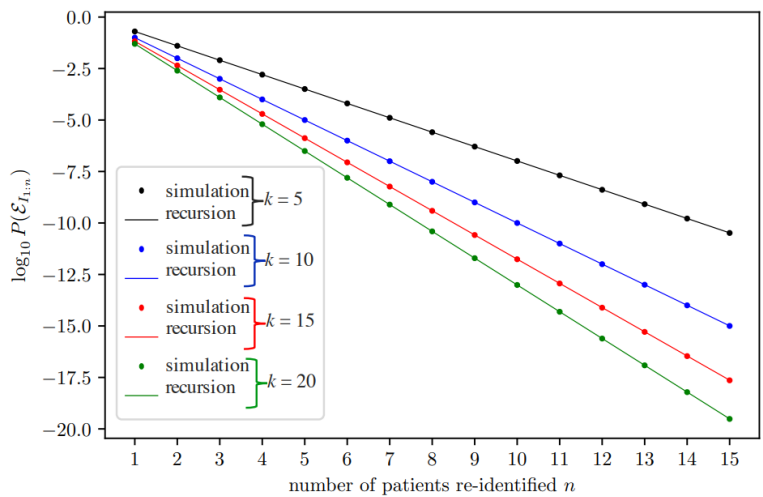

(a): Recursion vs simulation for multi-attack, $L = d = 10000$

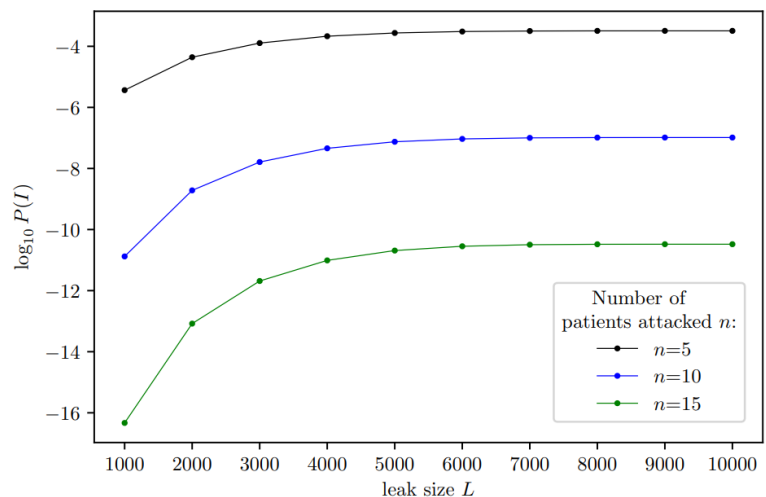

(b): Recursive probability of multi-patient attack, $k = 5, D = 10000$

Figure 4: Multiple patient re-identification probabilities. (a) Probability (log-scale) of re-identification for multiple patients for the special case $L = D = 10000$, $k = 5, 10, 15, 20$. The recursive solution agrees with the Monte Carlo simulation. (b) Recursive solution to the probability of re-identification for multiple patients (5, 10, 15) for varying leak size ($L = 1000, \ldots, D$) and an initial homogeneous equivalence class size ($k = 5$).

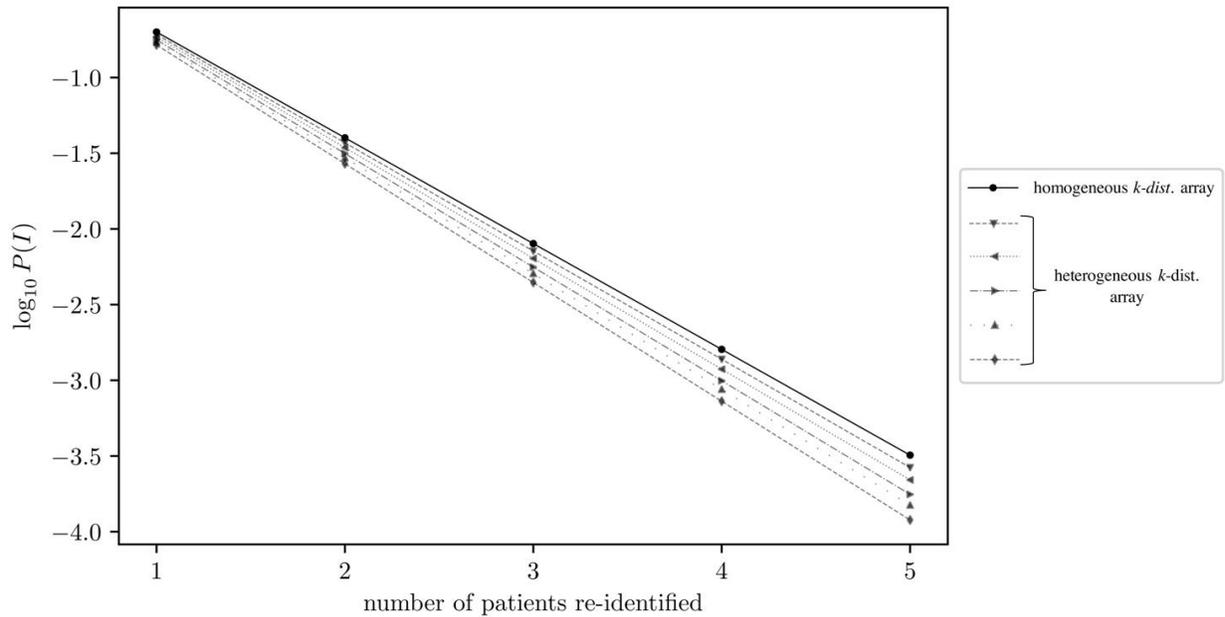

*Figure 5: Homogeneous vs heterogeneous state array for multi-patient attack, $L = D = 10000, k = 5$. The re-identification probabilities for a multi-patient attack using both a homogeneous (solid line) and multiple instances of a heterogeneous k-size distribution state array (dotted lines)*

## 4. DISCUSSION

As expected, the probability of re-identification increases with the size of the leak and has an inverse relationship with the $k$-class size. In addition, by quantifying the relationship between $k$ and the re-identification probability, we observed that the loss of granularity and subsequent increase in information loss arising from increasing $k$ provides diminishing returns to privacy for a fixed leak size; the benefits of a reduced re-identification risk that comes with increasing $k$-size may not be worth the reduction in granularity of the data when one is considering benchmarking the re-identification probability of a $k$-anonymised dataset on a specific leak size.

Moreover, our recursive approach provides a way of calculating the probability given any state of the system, parameterised by the $k$-size distribution, leak-size and total size. We have the ability to choose parameters of anonymisation to satisfy any re-identification probability threshold, and as our assumptions are conservative, the re-identification probability will always form an upper bound. Furthermore, the $k$-size distribution of an anonymised dataset might not be uniform. As seen here, the uniform $k$-size case results in a re-identification probability that is greater than any solution obtained using a distribution of equivalence class sizes with a minimum class size equivalent to the uniform case.

## 5. Limitations and further work

Our work is limited by conservative assumptions that may overestimate the reidentification probability in a $k$-anonymised dataset, namely that (1) a leak includes the entire set of information of a patient held in the dataset, and (2) the adversary holds the full set of quasi-identifiers of the patient they set out to re-identify.

In addition, our analysis does not address the potential damage caused by a homogeneity attack, that occurs when all the sensitive information in an equivalence class are identical- the main criticism of $k$-anonymity.

Future work could address leaks of partial patient records and take into account partial knowledge of patient sensitive information by the adversary. In addition, we will investigate the effects of $k$-anonymisation parameters on the true and false positive re-identification rates.

## 6. CONCLUSION

Our work provides a robust framework to quantify the effect of the $k$-anonymisation parameters and number of leaked records on multi-patient re-identification probability under the light of a re-identification attack due to a malicious data leak. This framework can guide regulators to setting appropriate bounds on the equivalence class size given an acceptable re-identification probability. By assessing the re-identification risks associated with data anonymisation, we hope this work will guide the adoption of safer anonymisation measures and thus safeguard privacy by allowing for a realistic evaluation of risk. Furthermore, more secure data protection measures will encourage the use of more datasets for research.

# REFERENCES

[1] S. C. Denaxas and K. I. Morley, "Big biomedical data and cardiovascular disease research: Opportunities and challenges," *European Heart Journal-Quality of Care and Clinical Outcomes*, vol. 1, no. 1, pp. 9–16, 2015.

[2] A. Baker, "Crossing the quality chasm: A new health system for the 21st century," *BMJ: British Medical Journal*, vol. 323, no. 7322, p. 1192, 2001.

[3] E. C. Lau, F. S. Mowat, M. A. Kelsh et al., "Use of electronic medical records (emr) for oncology outcomes research: Assessing the comparability of emr information to patient registry and health claims data," Clinical epidemiology, vol. 3, p. 259, 2011.

[4] G. Kanas, L. Morimoto, F. Mowat, et al., "Use of electronic medical records in oncology outcomes research," ClinicoEconomics and outcomes research: CEOR, vol. 2, p. 1, 2010.

[5] P. B. Jensen, L. J. Jensen, and S. Brunak, "Mining electronic health records: Towards better research applications and clinical care," Nature Reviews Genetics, vol. 13, no. 6, p. 395, 2012.

[6] L. Yao, Y. Zhang, Y. Li, et al. "Electronic health records: Implications for drug discovery," *Drug discovery today*, vol. 16, nos. 13-14, pp. 594–599, 2011.

[7] P. P. Yu, "Knowledge bases, clinical decision support systems, and rapid learning in oncology," *Journal of oncology practice, vol. 11, no. 2, pp. e206–e211, 2015.*

[8] *A. Rajkomar, E. Oren, K. Chen, et al. "Scalable and accurate deep learning with electronic health records," npj Digital Medicine, vol. 1, no. 1, p. 18, 2018.*

[9] M. R. Cowie, J. I. Blomster, L. H. Curtis, et al., "Electronic health records to facilitate clinical research," *Clinical Research in Cardiology, vol. 106, no. 1, pp. 1–9, 2017.*

[10] H. Hodson, "Revealed: Google AI has access to huge haul of NHS patient data," *New Scientist*, vol. 29, 2016.

[11] C. Cadwalladr and E. Graham-Harrison, "Revealed: 50 million facebook profiles harvested for cambridge analytica in major data breach," *The Guardian*, vol. 17, p. 2018, 2018.

[12] "Council of European Union. Regulation (EU) 2016/679." *Off. J. Eur. Union L*, vol. 119, pp. 1–88 (2016).

[13] M. A. Rothstein, "Is deidentification sufficient to protect health privacy in research?" *The American Journal of Bioethics*, vol. 10, no. 9, pp. 3–11, 2010.

[14] "Regulation (EU) 2016/679 of the European Parliament and of the Council of 27 April 2016 on the protection of natural persons with regard to the processing of personal data and on the free movement of such data, and repealing Directive 95/46/EC (General Data Protection Regulation)." [Online]. Available: http://data.europa.eu/eli/reg/2016/679/2016-05-04.

[15] 104th United States Congress, "The Health Insurance Portability and Accountability Act of 1996." 1996.

[16] K. El Emam, *Guide to the de-identification of personal health information*. Auerbach Publications, 2013.

[17] K. El Emam, S. Rodgers, and B. Malin, "Anonymising and sharing individual patient data," *BMJ* vol. 350, p. h1139, 2015.

[18] *Anonymisation:Managing data protection risk code of practice*. Information Commissioner's Office, 2012.

[19] A. Gkoulalas-Divanis, G. Loukides, and J. Sun, "Publishing data from electronic health records while preserving privacy: A survey of algorithms," *Journal of biomedical informatics*, vol. 50, pp. 4–19, 2014.

[20] L. Sweeney, "K-anonymity: A model for protecting privacy," *International Journal of Uncertainty, Fuzziness and Knowledge-Based Systems*, vol. 10, no. 5, pp. 557–570, 2002.

[21] K. El Emam, F. K. Dankar, R. Issa, et al., "A globally optimal k-anonymity method for the de-identification of health data," Journal of the American Medical Informatics Association, vol. 16, no. 5, pp. 670–682, 2009.

[22] K. El Emam and F. K. Dankar, "Protecting privacy using k-anonymity," *Journal of the American Medical Informatics Association*, vol. 15, no. 5, pp. 627–637, 2008.

[23] J. Domingo-Ferrer and V. Torra, "Disclosure risk assessment in statistical microdata protection via advanced record linkage," *Statistics and Computing*, vol. 13, no. 4, pp. 343–354, 2003.

[24] W. E. Winkler, "Matching and record linkage," *Business survey methods*, vol. 1, pp. 355–384, 1995.

[25] J. F. Robison-Cox, "A record linkage approach to imputation of missing data: Analyzing tag retention in a tag: Recapture experiment," *Journal of Agricultural, Biological, and Environmental Statistics*, pp. 48–61, 1998.

[26] W. E. Winkler, "Overview of record linkage and current research directions," in *Bureau of the census*, 2006.

[27] F. K. Dankar, K. El Emam, A. Neisa, and T. Roffey, "Estimating the re-identification risk of clinical data sets," *BMC medical informatics and decision making*, vol. 12, no. 1, p. 66, 2012.

[28] J. G. Bethlehem, W. J. Keller, and J. Pannekoek, "Disclosure control of microdata," *Journal of the American Statistical Association*, vol. 85, no. 409, pp. 38–45, 1990.

[29] L. Sweeney, "Uniqueness of simple demographics in the us population," *LIDAP-WP4, 2000*, 2000.

[30] K. El Emam, A. Brown, and P. AbdelMalik, "Evaluating predictors of geographic area population size cut-offs to manage re-identification risk," *Journal of the American Medical Informatics Association*, vol. 16, no. 2, pp. 256–266, 2009.

[31] P. Golle, "Revisiting the uniqueness of simple demographics in the us population," in *Proceedings of the 5th acm workshop on privacy in electronic society*, 2006, pp. 77–80.

[32] L. Rocher, J. M. Hendrickx, and Y.-A. de Montjoye, "Estimating the success of re-identifications in incomplete datasets using generative models," *Nature communications*, vol. 10, no. 1, p. 3069, 2019.

[33] J. Domingo-Ferrer, S. Ricci, and J. Soria-Comas, "Disclosure risk assessment via record linkage by a maximum-knowledge attacker," in *Privacy, security and trust (pst), 2015 13th annual conference on*, 2015, pp. 28–35.

[34] P. Samarati, "Protecting respondents identities in microdata release," *IEEE transactions on Knowledge and Data Engineering*, vol. 13, no. 6, pp. 1010–1027, 2001.

[35] G. Loukides, J. C. Denny, and B. Malin, "The disclosure of diagnosis codes can breach research participants' privacy," *Journal of the American Medical Informatics Association*, vol. 17, no. 3, pp. 322–327, 2010.

[36] A. Machanavajjhala, J. Gehrke, D. Kifer, and M. Venkitasubramaniam, "L-diversity: Privacy beyond k-anonymity," in *22nd international conference on data engineering (icde'06)*, 2006, pp. 24–24.

[37] Privitar, "Privitar publisher." https://www.privitar.com/publisher, 2019.

[38] Aircloak, "Aircloak insights." https://aircloak.com/, 2019.

[39] OpenAIRE2020, "Amnesia." https://amnesia.openaire.eu/index.html, 2019.

[40] "ARX - powerful data anonymization." http://arx.deidentifier.org/, 2019.